\documentclass[lettersize,journal]{IEEEtran}
\usepackage{amsmath,amsfonts}
\usepackage{algorithmic}
\usepackage{algorithm}
\usepackage{array}

\usepackage{textcomp}
\usepackage{stfloats}
\usepackage{url}
\usepackage{verbatim}
\usepackage{graphicx}
\usepackage{cite}
\usepackage{orcidlink}

\usepackage{subfigure}
\usepackage{multirow}
\usepackage{booktabs}
\usepackage{color}

\begin{document}

\title{LMAgent: A Large-scale Multimodal Agents Society for Multi-user Simulation}

\author{
Yijun Liu \orcidlink{0009-0006-7417-8399}, Wu Liu \orcidlink{0000-0003-1633-7575},~\IEEEmembership{Senior Member,~IEEE}, Xiaoyan Gu \orcidlink{0000-0003-0673-0058}, Xiaodong He \orcidlink{0000-0002-9463-9168},~\IEEEmembership{Fellow,~IEEE}, \\ Yong Rui \orcidlink{0000-0002-9142-5914},~\IEEEmembership{Fellow,~IEEE} and Yongdong Zhang \orcidlink{0000-0002-1151-1792},~\IEEEmembership{Fellow,~IEEE}

\thanks{Manuscript received XXX. This work was supported in part by xxx. Recommended for acceptance by xxx. (Corresponding author: Wu liu.)
}
\thanks{Yijun Liu is with the School of Information Science and Technology, University of Science and Technology of China; the Institute of Information Engineering, Chinese Academy of Sciences; the School of Cyber Security, University of the Chinese Academy of Sciences; Key Laboratory of Cyberspace Security Defense, Beijing, China, Beijing 100093, China. (e-mail: liuyijun@iie.ac.cn).}
\thanks{Wu liu, Yongdong Zhang are with the School of Information Science and Technology, University of Science and Technology of China, Hefei 230022, China. (e-mail: \{liuwu, zhyd73\}@ustc.edu.cn).}
\thanks{Xiaodong He is with JD AI Research, Beijing 100176, China (e-mail: xiaodong.he@jd.com).}
\thanks{Xiaoyan Gu is with the Institute of Information Engineering, Chinese Academy of Sciences; the School of Cyber Security, University of the Chinese Academy of Sciences; Key Laboratory of Cyberspace Security Defense, Beijing, China, Beijing 100093, China. (e-mail: guxiaoyann@iie.ac.cn).}
\thanks{Yong Rui is with Lenovo Research, Beijing 100094, China.  (e-mail: yongrui@lenovo.com).}
}

\markboth{Journal of \LaTeX\ Class Files,~Vol.~14, No.~8, August~2021}%
{Shell \MakeLowercase{\textit{et al.}}: A Sample Article Using IEEEtran.cls for IEEE Journals}

\maketitle

\begin{abstract}
The believable simulation of multi-user behavior is crucial for understanding complex social systems. 
Recently, large language models (LLMs)-based AI agents have made significant progress, enabling them to achieve human-like intelligence across various tasks.
However, real human societies are often dynamic and complex, involving numerous individuals engaging in multimodal interactions.
In this paper, taking e-commerce scenarios as an example, we present LMAgent, a very large-scale and multimodal agents society based on multimodal LLMs.
In LMAgent, besides freely chatting with friends, the agents can autonomously browse, purchase, and review products, even perform live streaming e-commerce.
To simulate this complex system, we introduce a self-consistency prompting mechanism to augment agents' multimodal capabilities, resulting in significantly improved decision-making performance over the existing multi-agent system.
Moreover, we propose a fast memory mechanism combined with the small-world model to enhance system efficiency, which supports more than 10,000 agent simulations in a society.
Experiments on agents' behavior
show that these agents achieve comparable performance to humans in behavioral indicators.
Furthermore, compared with the existing LLMs-based multi-agent system, more different and valuable phenomena are exhibited, such as herd behavior, which demonstrates the potential of LMAgent in credible large-scale social behavior simulations.
\end{abstract}


\section{Introduction}
The believable simulation of multi-user behavior has long been a pivotal challenge in artificial intelligence (AI), with AI agents regarded as promising tools for achieving this pursuit.
AI agents are artificial entities capable of perceiving the environment, making decisions, and taking action \cite{DBLP:journals/tmm/LanWWRZ24}.
By endowing AI agents with knowledge bases, objectives, and behavior capabilities, they can act consistently with their past experiences and react believably to their surroundings.
Furthermore, a system composed of multiple AI agents can construct highly complex sandbox environments. 
Such simulations are not only able to generate intricate interaction patterns, but can also be utilized to predict and analyze events that may occur in the real world \cite{park2023generative}. 

\begin{figure}
  \centering
    \subfigure[Multi-agent System]{\includegraphics[width=0.48\linewidth]{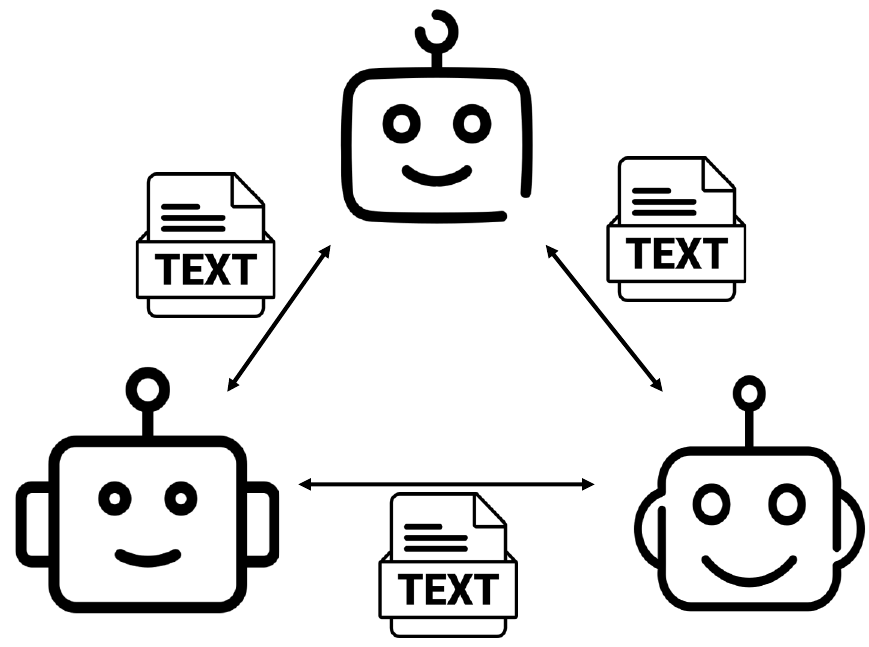}}
    \subfigure[LMAgent Society]{\includegraphics[width=0.5\linewidth]{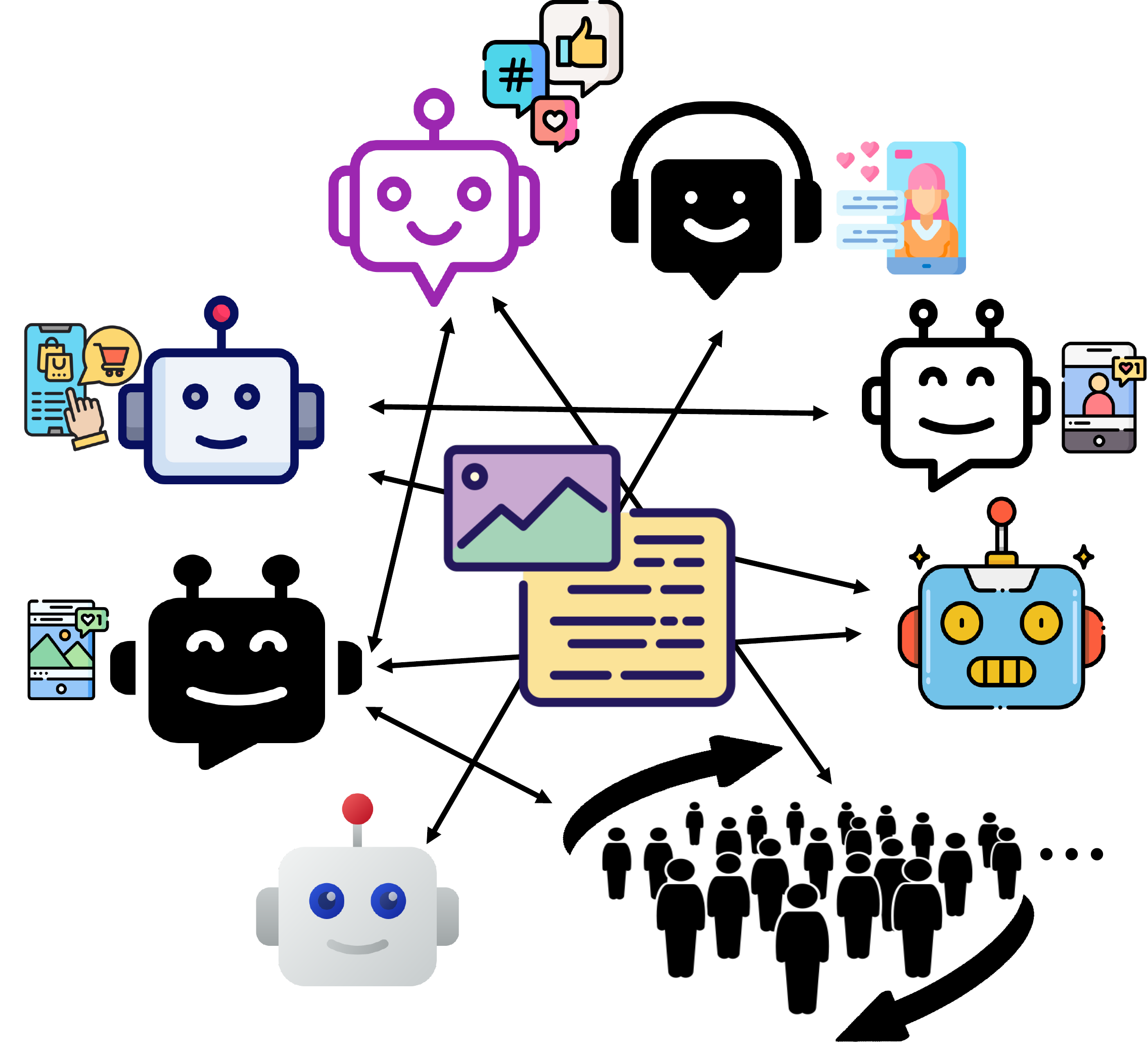}}
  \caption{(a) The existing Multi-agent System is driven by text-based LLMs, enabling textual interactions among multiple agents. (b) Our LMAgent is driven by multimodal LLMs, involving a society composed of ten thousand-scale agents and their multimodal interactions.}
  \label{fig:demo}
\end{figure}

Recently, large language models (LLMs) have achieved remarkable success and demonstrated significant potential in attaining human-like intelligence \cite{openai2023gpt}. 
LLMs exhibit strong capabilities in knowledge acquisition, command comprehension, generalization, planning, and reasoning \cite{wang2023describe}.
An increasing number of researchers are leveraging LLMs as central controllers to create AI agents endowed with human-like decision-making abilities \cite{DBLP:conf/acl/QianDLLXWC0CCL024,wang2023large,zhou2023sotopia}.
Various proof-of-concept AI agents such as GenerativeAgent \cite{park2023generative}, AgentVerse \cite{horton2023large}, and ChatDEV \cite{qian2023communicative} have showcased their application potential in respective domains. 
Moreover, to better simulate human cognitive mechanisms, recent studies have granted AI agents abilities in memory management \cite{lin2023swiftsage}, tool usage \cite{schick2023toolformer}, and task planning \cite{wang2023describe}. 
These advancements allow AI agents to make more effective decisions and accomplish tasks with unprecedented levels of autonomy.

However, while real human societies are inherently dynamic and intricate, explorations into large-scale simulations of online user behavior utilizing LLMs remain limited.
As shown in Figure~\ref{fig:demo}(a), most existing LLMs-based multi-agent systems only consider interactions among \textbf{a few agents} in the \textbf{text modality}, which overlooks the complexity of multimodal interactions in real-world settings. 
These limitations present the following challenges: \textbf{(1) How to integrate multimodal information and enhance the agents' multimodal analytical capabilities to accurately simulate user behaviors}; \textbf{(2) How to improve the operational efficiency of LLM-based agents to enable large-scale user behavior simulation.}

To handle these challenges, this paper introduces \textbf{LMAgent}: a very \textbf{large-scale} and \textbf{multimodal} agents society based on multimodal LLMs, as illustrated in Figure~\ref{fig:demo}(b).
To enhance the agents' multimodal analytical capabilities, we propose a self-consistency prompting mechanism that dynamically generates multimodal prompts through chain-of-thought reasoning.
This mechanism can significantly improve the consistency of the agent's decision-making in complex multimodal scenarios, thereby enhancing the simulation performance compared to text-only-based agents.
Given that large-scale user simulations with LLMs are computationally expensive, we introduce a fast memory mechanism that limits multimodal LLM calls to complex behaviors, reducing system load and improving efficiency by approximately 40\%.
The agent society is initialized using a small-world network model, in line with the six-degrees-of-separation theory \cite{milgram1967small}, which enhances communication efficiency among agents and more closely aligns with the real world. 

Through continuous evolution, this virtual agents society can even \textbf{exhibit emergent behaviors}, such as herd behavior, where they concentrate on purchasing certain products by the group acting, even if they do not need or like them.
Remarkably, our large-scale consumer simulations also produce co-purchase patterns that exhibit a \textbf{striking resemblance to real-world user data}, demonstrating its ability to replicate authentic consumer behavior.

To evaluate LMAgent, we conducted extensive experiments on the agent's behavior. 
We primarily assessed:
1) the user purchase behavior simulation capability of the agent in the sandbox environment;
2) the comparison between agent and human behavior and the influence of social factors on them;
3) the large-scale simulations of consumer behavior in e-commerce scenarios.
In summary, the main contributions of this paper are as follows:
\begin{itemize}
\item We propose a very large-scale and multimodal agents society, LMAgent. Based on this system, we conducted complex real-world user behavior simulations and achieved a comparable imitative ability as real humans.
\item Through self-consistency prompting, we introduce multimodal agents. Extensive experiments on agents' behavior and user purchase behavior simulation demonstrate that their performance significantly surpasses that of the existing LLMs-based multi-agent systems.
\item To enhance system efficiency, we propose a fast memory mechanism combined with the small-world model to support more than 10,000 agent simulations of agents society. As one of the biggest Agents Society Sandbox based on LLMs, it accurately captures large-scale user co-purchase patterns and even exhibits emergent behaviors, which demonstrates LMAgent's potential in credible large-scale social behavior simulations.
\end{itemize}

\begin{figure*}[!t]
  \centering
  \includegraphics[width=\linewidth]{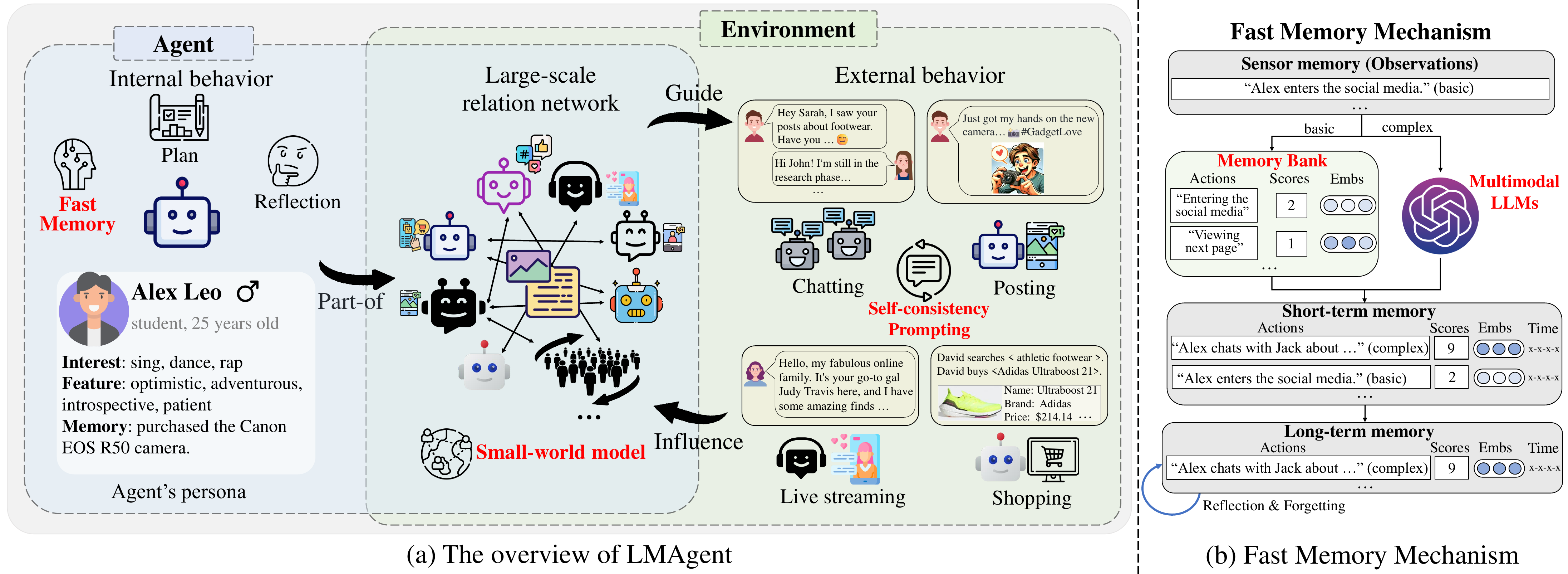}
  \caption{The overview of LMAgent. In this sandbox environment, each agent has its own memory and persona, it can set goals and reflect based on their memory. 
  From an external behavior perspective, agents can freely engage in multimodal social and shopping behaviors. 
  Their internal behavior can guide their external behavior, which in turn influences their internal behavior.
  We use the small-world model to initialize the society's relation network to more closely resemble real-world social networks.
  }
  \label{fig:overview}
\end{figure*}

\section{Related Work}
\subsection{Believable Proxies of User Behavior}
The credible simulation of user behavior has always been the key to studying complex social systems 
\cite{chuang2024simulating}.
The design of believable proxies creates an illusion of human behavior, allowing them to make decisions and take actions according to their will
\cite{park2023generative}.
Over the past few decades, various methods \cite{mnih2013playing,DBLP:journals/tmm/LinYFLYXC24,DBLP:journals/tmm/SlavicBCMR22} have
been proposed to create credible agents. 
Rule-based methods such as finite-state machines \cite{masek2018discovering} and behavior trees \cite{colledanchise2018learning} provide a direct approach to creating simple agents, which remains the primary method today for handling basic social interactions.
Nevertheless, manually crafting response rules for all behaviors is unsustainable. 
Some scholars have proposed using reinforcement learning methods to automate decision-making \cite{DBLP:journals/tmm/LvFNDJZXX23,DBLP:journals/tmm/NieWLCWJLL24,DBLP:journals/tmm/WangC23}. 
RecSim \cite{ie2019recsim} uses reinforcement learning to simulate users' continuous behavior for interactive recommendations. 
AlphaStar \cite{arulkumaran2019alphastar} and DQN \cite{mnih2013playing} enable agents to autonomously learn in unknown environments, allowing them to achieve impressive performance in some decision-making games. 
However, their success mainly stems from easily defined reward functions, which can be optimized by learning algorithms. 
In an open agents society, there is often no clear reward function for learning, and the decision-making process for agent behavior becomes highly complex.

In our work, we employ multimodal LLMs as central controllers to design multimodal agents. 
These agents possess human-like decision-making capabilities, enabling them to make rational decisions in complex and dynamic environments for simulating credible user behavior.

\subsection{LLMs-based Agent System}
LLMs contain a wealth of knowledge about the world and can generate human-like responses based on social contexts \cite{openai2023gpt}. 
LLM-based agent systems refer to systems that utilize LLMs as engines to drive multiple agents to make various behavioral decisions \cite{park2023generative}.
In this system, each agent holds its own knowledge base, goals, and abilities. 
They can interact and collaborate to enhance the system's ability to handle complex tasks and dynamic environments \cite{chen2023agentverse,lin2023swiftsage,wang2023describe}.
Recently, there has been significant progress in LLM-based agent systems.
\cite{aher2023using} conducted preliminary tests and found that LLMs possess the ability to replicate some classic experiments in economics, psycholinguistics, and social psychology. 
\cite{horton2023large} utilized LLMs-based agents to substitute human participants, endowing the agents with talents, backgrounds, and preferences, and prompting them to simulate economic behavior.
The results with these LLM-empowered agents are similar in quality to those of human experiments.
\cite{chuang2024simulating} simulated human opinion dynamics based on multi-agent systems, revealing a strong inherent bias in LLM agents towards producing accurate information.
In addition, researchers have also employed LLMs to construct other agent systems for simulation or to improve work efficiency. 
For example, collaborative software development \cite{qian2023communicative,hong2023metagpt,DBLP:conf/acl/QianDLLXWC0CCL024}, social simulation \cite{wang2023large,park2022social,zhou2023sotopia} and game
playing \cite{xu2023exploring,hua2023war,gong2024mindagent}.

Different from the previous work, our work aims to construct a large-scale agents society for more realistic and credible multi-user behavior simulation. 
Meanwhile, we also endow these agents with multimodal capabilities, enabling them to make decisions and take actions like real humans.

\section{Method}
This section outlines the multimodal agent architecture and the sandbox environment within LMAgent. 
To concretize the behavior of agents, we instantiate agents as consumers in e-commerce scenarios. 
However, LMAgent is \textbf{versatile}. By expanding the types of agent behaviors, it can easily adapt to other scenarios.

\subsection{Multimodal Agent Architecture}
Sociologists believe that when analyzing individuals, two dimensions should be considered: external and internal behaviors.
The external dimension relates to the individual's observable actions, while the internal dimension involves personality, values, and emotions. 
As illustrated in Figure~\ref{fig:overview}, LMAgent offers perspectives on both the internal and external behaviors of agents.
These behaviors are all powered by multimodal LLMs\footnote{The LLM we use is ChatGPT (version: gpt-4-1106-preview and gpt-4-vision-preview): \url{https://openai.com/gpt-4}}.
Internally, each agent possesses distinct modules for persona and memory. 
The persona module shapes character traits for the agent, allowing for more personalized behavior. 
The memory module controls the writing and forgetting of the agent's memories. 
Externally, each agent can freely engage in social or shopping activities, which involve both vision and text-modal information.
It's noteworthy that to improve the multimodal analysis ability of the agents, we propose a self-consistency prompting mechanism to dynamically construct prompts for external behaviors.
To enhance system efficiency, we have designed a fast memory module for agents to increase their action speed.
Next, we will elaborate on these behaviors in detail.

\subsection{Internal Behavior}
\subsubsection{Persona}
The persona is an important concept in social simulation. 
To endow each agent with distinct social backgrounds and personal characteristics, we have introduced a set of generic attributes to construct the agent's persona, including name, gender, age, occupation, personal traits,
purchasing preferences
and behavioral tendencies.
The name, occupation, and personal traits are randomly assigned by LLM; age is randomly assigned and the population follows a truncated normal distribution; preferences and behavioral tendencies are inferred using LLM based on the above information.

\subsubsection{Fast Memory}
Memory mechanisms are integral to the agent's cognition of virtual environments and are key to the design of agents. 
In a large-scale and multimodal agents society, a slow and inefficient memory mechanism can significantly increase the economic and time costs of the system. 
To address this issue, we have designed a fast memory mechanism to enhance the system's efficiency.
As illustrated in Fig~\ref{fig:overview}, the design of fast memory is built on \cite{wang2023large} and optimized specifically for \textbf{large-scale and multimodal} scenarios.
It aligns with the advancements in cognitive neuroscience \cite{atkinson1968human}, encompassing sensor, short-term, and long-term memory. A memory bank is proposed to enhance the system's efficiency.

\textbf{Sensor memory} is used to process the currently observed information $o_i$ in time $i$.  
It records all details of $o_i$ but then immediately forgets them, with key information being condensed into more informative, concise sentences $c^s_i$, then stored in short-term memory, denoted as:
\begin{equation}
    c^s_i = f_c(o_i),
\end{equation}
where $o_i$ \textbf{can be text and/or images}, $f_c$ is the prompt function to guide the LLMs in information compression.
This process eliminates irrelevant and unimportant content while also compressing information to save space and enhance operational efficiency. 
\textbf{Short-term memory} stores compressed sensor memories as formatted memory. Each agent $a_i$ has an independent short-term memory base $\mathcal{M}^s = \{m^s_1, m^s_2, ..., m^s_N\}$, where $m^s_i = <c^s_i, e_i, I_i, t_i>$ is the formatted memory record.
$e_i$ is the embedding
\footnote{The embedding model we use is text-embedding-ada-002 provided by OpenAI: https://platform.openai.com/docs.} 
of $c^s_i$; $t_i$ and $I_i$ in $m^s_i$ are the time stamp and the importance score. 
We use LLMs for scoring the memory importance $I_i$ to distinguish between mundane and core memory:
\begin{equation}
    I_i = f_r(c^s_i),
\end{equation}
where $f_r$ is the prompt function to guide the LLMs in rating memory.

\textbf{Long-term memory} stores information of more importance, which can be retrieved before an agent's actions. 
We use the cosine similarity of $e_i$ and $e_j$ to measure the distance between $m^s_i$ and $m^s_j$. 
When $K$ similar memories to $m^s_i$ appear, we denote $m^s_i$ as $m^l_i$ and store it in the long-term memory base $\mathcal{M}^l = \{m^l_1, m^l_2, ..., m^l_N\}$.
Moreover, cognitive neuroscience \cite{nairne2007adaptive} suggests that long-term memory exhibits varying probabilities of forgetting based on time and importance.
Intuitively, the older and less important memories are more likely to be forgotten.
Specifically, we use the following memory-forgetting formula:
\begin{equation}
    f\left(m^l_i\right)=1-\frac{\hat{t}_i+I_i}{2} * \max \left(I_i^\beta, \delta\right),
\end{equation}
where $\hat{t}_i$ is the recency scores normalized based on the span between the oldest and newest memories, with the oldest and newest memories scored as 0 and 1, and those in between scored proportionally.
$\beta$ is a hyper-parameter controlling the power function shape.
$\delta$ is a strength parameter determining when to consider the power function of $I_i^\beta$. 
In this formula, the older and less important memories are more likely to be forgotten, which is more rational and in line with cognitive neuroscience research \cite{nairne2007adaptive}.

\textbf{Memory bank} is the key to enhancing system efficiency.
The majority of what the agent observes is basic behaviors, such as entering social media, entering shopping malls, and so on. 
Compressing and scoring all observations using LLMs is inefficient, so we designed a memory bank $\mathcal{B} = \{m^c_i \mid i=1, 2, ..., N\}$ to cache the basic behavior information $m^c_i$ of the agent, where $m^c_i = <y_i, I_i, e_i>$, $y_i$ is the action types.
When it comes to basic behaviors, the agent can directly retrieve $I_i$ and $e_i$ from $\mathcal{B}$, thus eliminating the need to call the LLMs.
In our statistics, the basic behaviors of agents account for over 60\% of all actions. By employing the memory banks, we can yield an efficiency improvement of approximately 40\%.

\subsubsection{Planning and Reflection} 
Following \cite{park2023generative}, we incorporate planning and reflection to urge high-level thinking in the agent. 
Planning involves agents setting goals based on their characteristics and experiences, making their overall behavior logic more reasonable. 
Reflection, on the other hand, is thinking through existing memories to gain higher-level insights.
The steps include: 1) \textit{Generating the most salient questions based on the agent's recent experiences}. 2) \textit{Retrieving relevant memories to answer these questions and extracting high-level insights}. The reflection results are stored in long-term memory to guide the agent's subsequent behavior implicitly.


\subsection{External Behavior}

\subsubsection{Shopping and Social Interaction}
Shopping and social behaviors are the most critical behaviors in e-commerce scenarios. 
For shopping, we designed a comprehensive set of shopping actions, including: 
1) \textbf{Browsing}: observing the products recommended by the shopping system; 
2) \textbf{Searching}: searching for specified products;
3) \textbf{Paging}: viewing more recommended products;
4) \textbf{Viewing Details}: inquiring about detailed product information;
5) \textbf{Purchasing}: buying specified products. 
For social interactions, we also designed a set of common social actions, including:
1) \textbf{Chatting}: choose a friend for a conversation.
2) \textbf{Posting}: post message to all friends. 
3) \textbf{Live streaming}: a few ``superstar'' agents can perform live streaming to introduce and recommend products, which is of great significance to studying the live commerce phenomenon in the real world.
Notably, these behaviors involve multimodal interactions, allowing agents to directly convey visual information.

\subsubsection{Self-consistency Prompting}
Decision-making in e-commerce scenarios often requires consideration of multimodal information about users themselves and the environment, which is challenging for LLMs. 
Inspired by chain-of-thought \cite{wei2022chain}, we \textbf{divide the decision-making of these behaviors into two stages} and \textbf{dynamically construct self-consistent prompts} to improve the decision-making ability of agents.
For example, if we want to determine the agent's next shopping action $a$:
1) In the first stage, we focus on internal information, making summaries $\mathcal{P}_1$ based on personal characteristics $\mathcal{C}_i$ toward the last observation $o_i$ (such as: enter the shopping system):
\begin{equation}
    \mathcal{P}_1 = f_s(\mathcal{C}_i, o_i),
\end{equation}
where $f_s$ is the prompt function to guide the LLMs in making summaries. 
This emphasizes personal features to improve the self-consistency for the next action.
2) In the second stage, we incorporate $\mathcal{P}_1$ and the multimodal environmental information $\mathcal{E}$ (e.g. the product's images and description) to form the final prompts for inferring the agent's next action $a$.
\begin{equation}
    a = f_e(\mathcal{P}_1, \mathcal{E}),
\end{equation}
where $f_e$ is the prompt function to guide the LLMs in making decisions based on the environment $\mathcal{E}$ and user characteristics $\mathcal{P}_1$.
By decoupling the task in this manner, LLMs need only focus on a portion of the decision-making process in each stage, thereby enhancing the credibility of the decision as well as its self-consistency.

\begin{figure}[!t]
  \centering
  \includegraphics[width=\linewidth]{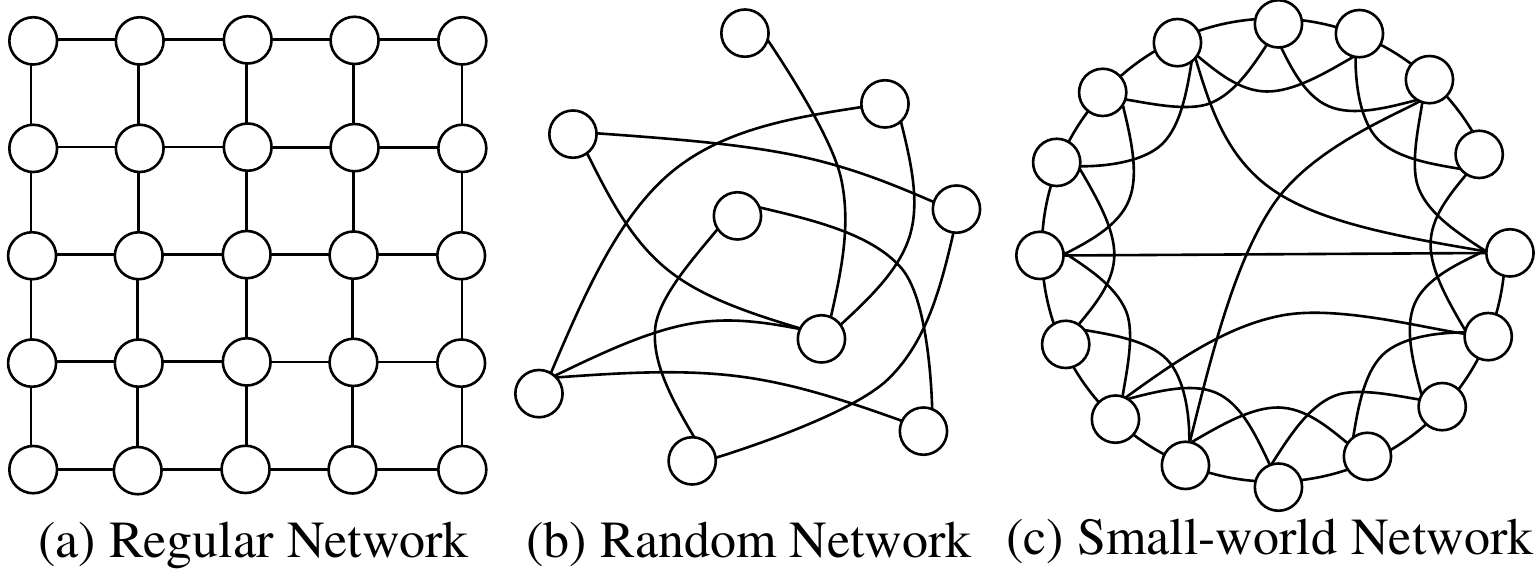}
  \caption{Diagram of different network structures.
  }
  \label{fig:net}
\end{figure}

\begin{algorithm}[tb]
\caption{Small World Topology Network Construction}
\label{alg:sw}
\textbf{Input:} Number of agents $N$, average number of friends for all agents $k$, rewiring probability $p$\\
\textbf{Output:} \makebox[0.1\textwidth][l]{Small-world network graph $\mathcal{G} = (\mathcal{V}, \mathcal{A})$}
\begin{algorithmic}[1]
\STATE Initialize $\mathcal{A}=\mathbf{0}_{N \times N}$ and $\mathcal{V} = \{v_i \mid i=1, 2, ..., N\}$

\FOR{each node $i = 1$ to $N$}
    \FOR{$j = 1$ to $k/2$}
        \STATE $\mathcal{A}[i, (i+j) \mod N] \gets 1$
        \STATE $\mathcal{A}[(i+j) \mod N, i] \gets 1$
    \ENDFOR
\ENDFOR
\FOR{each node $i = 1$ to $N$}
    \FOR{each neighbor $j$ such that $A[i, j] = 1$}
        \STATE Generate a random number $r$  uniformly from $[0,1)$
        \IF{$r < p$}
            \STATE Select a random node $m$ such that $m \neq i$ and $\mathcal{A}[i, m] = 0$
            \STATE Update $\mathcal{A}[i, j] \leftarrow 0$ and $\mathcal{A}[j, i] \leftarrow 0$
            \STATE Update $\mathcal{A}[i, m] \leftarrow 1$ and $\mathcal{A}[m, i] \leftarrow 1$
        \ENDIF
    \ENDFOR
\ENDFOR
\STATE \textbf{return} $\mathcal{G}$
\end{algorithmic}
\end{algorithm}
\subsection{Sandbox Environment}
\label{sec:se}
\subsubsection{Small-world Topology Networks}
To improve communication efficiency between agents and construct large-scale social networks more aligned with the real world, we employed the \textbf{small-world model} \cite{watts1998collective} to initialize agents' social networks.
As shown in Figure~\ref{fig:net} (c), the small-world network has \textbf{a higher clustering coefficient and a shorter average path length} than other networks.
This reflects the presence of local clustering and rapid information spread, similar to real-world networks.
The construction of the small world topology network is shown in Algorithm~\ref{alg:sw}.

Formally, given the relation graph $\mathcal{G} = (\mathcal{V}, \mathcal{A})$, we first initialize it with:
\begin{equation}
    \mathcal{A}=\mathbf{0}_{N \times N}, 
\end{equation}
\begin{equation}
    \mathcal{V} = \{v_i \mid i=1, 2, ..., N\},
\end{equation}
where $\mathcal{V}$ represents the set of nodes (agents). $\mathcal{A}$ denotes the adjacency matrix of the agent's relations.

Then we arrange $N$ nodes in a one-dimensional lattice with periodic boundary conditions (i.e., forming a ring). Each node $i$ is connected to its $k$ nearest neighbors, $\frac{k}{2}$ on each side, ensuring a \textbf{high clustering coefficient}:
\begin{equation}
    \mathcal{A}_{i j}= 
    \begin{cases}1 & \text { if } 1 \leq|i-j| \leq \frac{k}{2}\\ 1 & \text { if } N-\frac{k}{2} \leq|i-j| \leq N-1 \\ 
    0 & \text { otherwise }\end{cases}.
\end{equation}

After that, we rewrite all edges with a probability $p$ to introduce long-range connections, \textbf{significantly reducing the average path length of $\mathcal{G}$}.
Specifically, for each node $i \in \mathcal{V}$, for each neighbor $j$ of $i$, we generate a random number $r$  uniformly from $[0,1)$, if $r \leq p$, we rewrite the edge $(i,j)$:
\begin{equation}
    \mathcal{A}[i, j] \leftarrow 0, \mathcal{A}[j, i] \leftarrow 0,
\end{equation}
\begin{equation}
    \mathcal{A}[i, j^{\prime}] \leftarrow 1, \mathcal{A}[j^{\prime}, i] \leftarrow 1, j^{\prime} \in \mathcal{V} \backslash\{i\} \cup \mathcal{N}(i),
\end{equation}
where $\mathcal{N}(i)$ is the set of neighbors of $i$.

In this way, we can ensure that the $\mathcal{G}=(\mathcal{V}, \mathcal{A})$ roughly follows the \textbf{six-degree-of-separation space theory} \cite{milgram1967small} (where any two strangers are separated by no more than six people), and thus resemble realistic social networks. 
It can also improve the agents' communication efficiency, as information can be disseminated over greater distances with fewer connections.
The system's cost increases linearly with the number of agents $N$, the time complexity is $kN$, where $k$ is the average number of friends for all agents.

\begin{algorithm}[tb]
\caption{Multi-user Behavior Simulation}
\label{alg:sb}
\textbf{Input:} $N$ agent $\{\textbf{a}_i\}^N_{i=1}$ with their personas $\{\mathcal{C}_i\}^N_{i=1}$ and memories $\{\mathcal{M}_i=(\mathcal{M}^s_i,\mathcal{M}^l_i)\}^N_{i=1}$, external behavior set $\mathbb{A}$, time steps $T$, and the agents' relationship network $\mathcal{G}$\\
\textbf{Output:} \makebox[0.1\textwidth][l]{Multi-user simulation logs $\mathbb{L}$}
\begin{algorithmic}[1]
\STATE Initialize $\mathcal{G}$ to small-world network and assign the persona $\mathcal{C}_i$ to agent $\textbf{a}_i$

\FOR{each time step $t$ in $1$ to $T$}
    \FOR{each agent $\textbf{a}_i$, $i$ in $1$ to $N$}
        \STATE Planning and reflection (if necessary) based on $\mathcal{M}_i$ and $pro_i$, and store the results in $\mathcal{M}_i$.
        \STATE Select the next action $a$ from $\mathbb{A}$, based on $\mathcal{M}_i$, $pro_i$ by self-consistency prompting.
        \STATE Execute action $a$ according to the system process
        \STATE Manipulate $\mathcal{M}_i$ according to the fast memory mechanism and add logs to $\mathbb{L}$
    \ENDFOR
\ENDFOR
\STATE \textbf{return} $\mathbb{L}$
\end{algorithmic}
\end{algorithm}

\subsubsection{Multi-user Simulator}
As shown in Algorithm~\ref{alg:sb}, in this sandbox environment, agents take turns acting, freely engaging in multimodal social or shopping interactions. 
These behaviors can change their memory and affect other agents' behavior.
During system operation, we can pause and observe the status of the agents at any time, or continuously run the simulation to examine the evolution of social phenomena. 

\section{Experiments}
LMAgent is designed to construct a very large-scale and multimodal agents society for complex user behavior simulation.  
Since we use the e-commerce scenario as a concrete instance for LMAgent, our experiment mainly focuses on agents' \textbf{shopping} and \textbf{social} behaviors.
Furthermore, we conducted \textbf{large-scale consumer simulation} to analyze the group behavior generated by LMAgent, validating its potential in large-scale social simulations.

\subsection{Experimental Setup}
In our experiments, the LLM used is ChatGPT (version: gpt-4-1106-preview) accessed via OpenAI API calls \cite{openai2023gpt}.
The random seed is set to 1 for reproducibility, and the system is implemented based on LangChain \cite{Chase_LangChain_2022}.
The dataset in the virtual shopping system is initialized with the Amazon Review Dataset \cite{amazonReviewDataset} $\mathcal{D}_a$, which contains 233.1 million unique purchases and reviews information from over 20 million users.
It also includes detailed product information, such as product names, prices, images, etc, making it the largest and most comprehensive dataset in the recommendation field.

\subsection{User Purchase Behavior Evaluation}
\label{subsec:mrse}
To quantitatively evaluate the agent behavior in dynamic and complex scenarios,  we simulate user purchase behavior 
based on the
sandbox environment of LMAgent. 
In this task, each agent needs to make decisions on future purchases based on their purchase history and the multimodal information of the products.
Specifically, for each simulated user $u$, we initialize it using the real shopping history $H_u$ in the $\mathcal{D}_a$.
We retain the last $a$ items of $H_u$ as ground truth $T_u$ for evaluation, and utilize all other items to initialize the agent's persona.
When shopping, we combined $T_u$ with $b$ random items from the product database to form a recommendation list $R_u$.
We present the $R_u$ to the agent, and let it select $a$ items $S_u$ to compare with the $T_u$. 

\subsubsection{Evaluation Metrics}
To facilitate comparisons with existing works, we follow the metric $a@(a+b)$ proposed by \cite{wang2023large} to evaluate the purchase accuracy of agents, where $a$ and $(a+b)$ are the quantity of ground truth and recommended list products, respectively. 
Specifically, for a user $u$, let $S_u$ indicate the predicted set of purchased products, and $T_u$ denotes the ground truth. 
We then employ the following metric to assess the performance of different models:
\begin{equation}
    p = \sum_{u \in U}\frac{|T_u \cap S_u|}{|T_u|} \times 100\%,
\end{equation}
where $U$ is the set of all simulated users.
, and a larger $p$ indicates better performance. 
In our experiment, we set $a$ and $b$ to different values to provide a more comprehensive evaluation of our system.
Intuitively, selecting fewer products from a larger number of candidate products should be more difficult.
Moreover, due to the \textbf{large scale} of experiments and the \textbf{independence} of agents' actions, it inherently avoids randomness issues and can statistically significantly reflect the model's performance.

\begin{table}[!t]
\centering
\caption{The results of different models on user purchase simulation under various $a@(a+b)$ settings. \textbf{Best} (bold) are highlighted in each column.} 
\label{tab:res1}
\begin{tabular}{cccccc}
\toprule
\textbf{Model}                                                    & \textbf{1@6}    & \textbf{1@10}   & \textbf{3@6}    & \textbf{3@10}   & \textbf{AVG} \\
\midrule
Random                                                            & 16.00 & 11.20 & 51.07 & 28.67 & 26.74  \\
Embedding     \cite{mnih2007probabilistic}                                                   & 37.60 & 23.20 & 65.47 & 48.53 & 43.70  \\
\begin{tabular}[c]{@{}c@{}}Collaborative\\ Filtering \cite{ekstrand2011collaborative}\end{tabular} & 52.80 & 32.40 & 67.87 & 52.67 & 51.44  \\
\midrule
Recsim     \cite{ie2019recsim}                                                     & 48.40 & 43.60 & 75.33 & 57.73 & 56.27  \\
RecAgent  \cite{wang2023large}                                                     & 52.40 & 46.00 & 73.87 & 61.47 & 58.44  \\
\midrule
\textbf{LMAgent}                                                           & \textbf{70.40} & \textbf{63.60} & \textbf{82.67} & \textbf{75.47} & \textbf{73.04} \\
\bottomrule
\end{tabular}
\end{table}

\subsubsection{Results}
We compared LMAgent with some well-known recommendation algorithms and \textbf{achieved the state-of-art performance.}
The baseline models we use include Embedding \cite{mnih2007probabilistic}, Collaborative Filtering \cite{ekstrand2011collaborative}, and some multi-agent-based recommendation systems like Recsim \cite{ie2019recsim} and RecAgent \cite{wang2023large}. 
Among them, traditional methods can only consider purchase history and product information, whereas multi-agent-based approaches can incorporate additional user profiles or social information.
Table~\ref{tab:res1} presents the performance of different models in the user purchase behavior evaluation. 
As shown in the table, the LLM-based agent system consistently outperforms traditional recommendation algorithms across all experimental settings. On average, LMAgent achieves a performance improvement of approximately 29.34\% over the baseline methods. Notably, the improvement is even more pronounced in the more challenging 1@6 and 1@10 settings, reaching an average of 32.80\%.
These results underscore the importance of multimodal information and self-consistency prompting in e-commerce scenarios, marking LMAgent as a significant step forward in accurate user behavior simulation.

\subsection{Agent Behavior Analysis}
\label{sec:aba}
In this section, we broaden the scope of the evaluation, analyze the agent's behavior from different dimensions, and compare it with human behavior to verify the effectiveness of LMAgent in user behavior simulation.
\subsubsection{Data Collection}
In this experiment, we evaluate the logical coherence of agents by analyzing their \textbf{behavior chain} and \textbf{behavior content}. A behavior chain \textbf{is a sequence of actions} performed by the agent, such as: \textit{Accessing social media - Chatting with Mary Williams - Entering the shopping system}. Behavioral content refers to the \textbf{social outputs} generated by agents, including posts and chat history.
Data was collected from 1,000 agents running 10 rounds in LMAgent for analysis. Additionally, to benchmark human behavior, we gathered data from 50 volunteers who controlled 500 agents, allowing us to compare the agent behaviors against real-world human actions and validate LMAgent’s simulation accuracy.

\begin{table}[!t]
\centering
\caption{
Evaluation of behavior chains of agents and humans in the sandbox environment.
Each indicator is scored by \textbf{Humans (H)/GPT-4 (G)} on a scale of 1 to 5. \textbf{Best} is in bold, values close to human results (difference $\le$ 0.3) are in \color{blue}{blue}.}
\label{tab:res2}
\begin{tabular}{ccccccc}
\toprule
\multirow{2}{*}{\textbf{Dim}} & \multicolumn{2}{c}{\textbf{Random}}          & \multicolumn{2}{c}{\textbf{LMAgent}}         & \multicolumn{2}{c}{\textbf{Human}}           \\ \cline{2-7} 
                              & \multicolumn{1}{c|}{\textbf{H}} & \textbf{G} & \multicolumn{1}{c|}{\textbf{H}} & \textbf{G} & \multicolumn{1}{c|}{\textbf{H}} & \textbf{G} \\ \hline
Believability                 & \multicolumn{1}{c|}{2.70}       & 3.17       & \multicolumn{1}{c|}{4.24}       & 3.72       & \multicolumn{1}{c|}{\textbf{4.80}}       & 3.33       \\
Knowledge                     & \multicolumn{1}{c|}{2.75}       & 3.22       & \multicolumn{1}{c|}{\color{blue}{4.05}}       & 3.89       & \multicolumn{1}{c|}{\textbf{4.20}}       & 2.83       \\
Personalization               & \multicolumn{1}{c|}{2.68}       & 4.10       & \multicolumn{1}{c|}{4.20}       & 4.46       & \multicolumn{1}{c|}{\textbf{4.53}}       & 3.77       \\
Social Norms                  & \multicolumn{1}{c|}{4.33}       & 3.10       & \multicolumn{1}{c|}{\color{blue}{4.59}}       & 3.64       & \multicolumn{1}{c|}{\textbf{4.87}}       & 3.53       \\
Social Influence              & \multicolumn{1}{c|}{2.93}       & 3.83       & \multicolumn{1}{c|}{\color{blue}{4.43}}       & 4.11       & \multicolumn{1}{c|}{\textbf{4.60}}       & 3.67       \\ \hline
Average                       & \multicolumn{1}{c|}{3.08}       & 3.48       & \multicolumn{1}{c|}{\color{blue}{4.30}}       & 3.96       & \multicolumn{1}{c|}{\textbf{4.60}}       & 3.43       \\
\bottomrule
\end{tabular}
\end{table}
\subsubsection{Evaluation Metrics}
To thoroughly assess the agents' behavioral patterns, we developed a multidimensional evaluation framework informed by anthropology, psychology, and sociology. This framework encompasses seven distinct dimensions designed to capture the nuances of agent behavior. Each dimension is evaluated through human scoring on a 1–5 scale, providing qualitative insights alongside quantitative analysis. 

For the evaluation of behavior chains, we focus on five key dimensions:
\begin{itemize}
    \item \textbf{Believability} \cite{park2023generative}, which gauges the extent to which the agent’s actions appear plausible and authentic within the given context.
    \item \textbf{Knowledge} \cite{zhou2023sotopia}, assessing the agent's ability to demonstrate domain-relevant expertise and accurate information during interactions.
    \item \textbf{Personalization} \cite{roberts2007power}, which evaluates the agent's capacity to adapt its actions and responses to individual user preferences and characteristics.
    \item \textbf{Social Norms} \cite{yamin2019using}, which considers the agent’s alignment with established societal conventions and behavioral expectations.
    \item \textbf{Social Influence} \cite{coleman1986social}, capturing the degree to which the agent’s behavior is shaped by or influences the actions of other agents or external factors.
\end{itemize}

In addition, we assess behavioral content along two dimensions:
\begin{itemize}
    \item \textbf{Naturalness} \cite{sinclair1984naturalness}, reflecting the degree to which the content generated by the agent mirrors natural, human-like communication.
    \item \textbf{Expressiveness} \cite{chatterjee2010expressiveness}, which measures the ability of the agent to convey emotions, intent, and nuanced social signals through its content.
\end{itemize}
Here is an example of evaluating the Naturalness dimension:
\begin{figure}[!h]
  \centering
  \includegraphics[width=\linewidth]{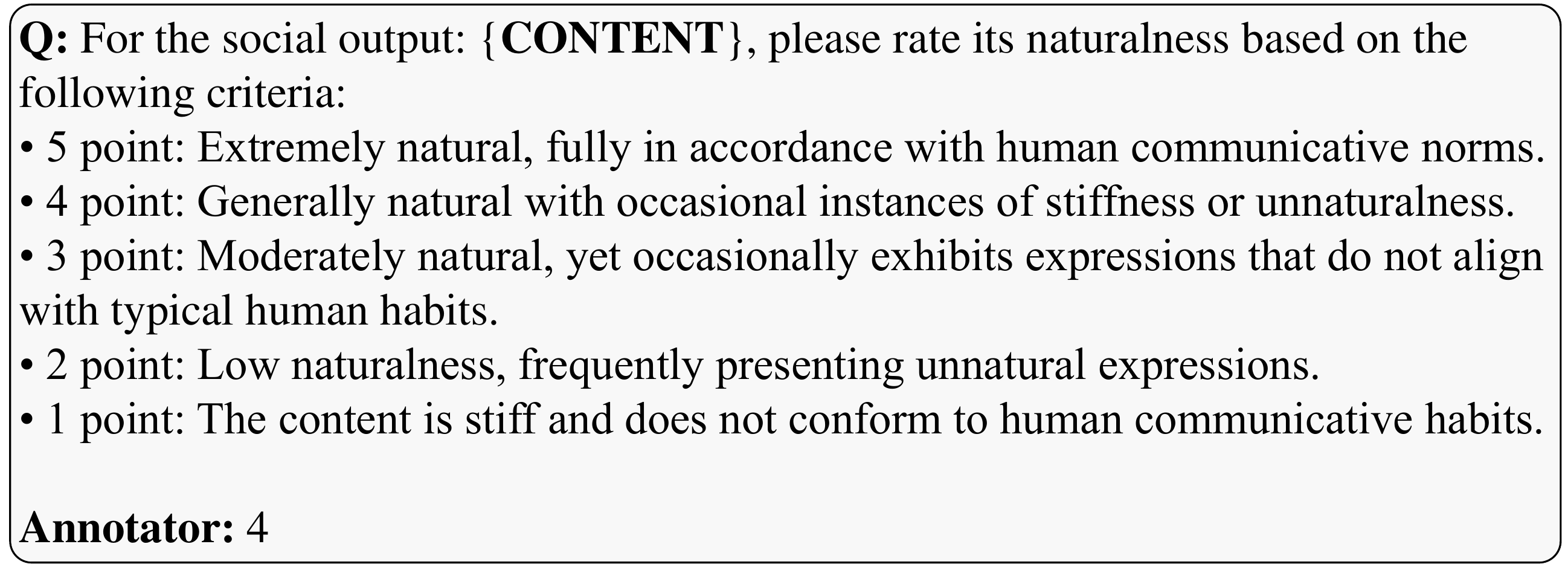}
\end{figure}

To ensure robust evaluation, 100 volunteers from a wide array of cultural and professional backgrounds were recruited to annotate the data. The level of agreement among annotators was found to be moderate, with an average Randolph $\kappa$ score \cite{randolph2005free} of 0.573, indicating a reasonable degree of consistency in judgment.
Additionally, following recent work \cite{zhou2023sotopia}, we utilize GPT-4’s scoring as an auxiliary reference for evaluation. 
However, we remind readers that LLMs may exhibit issues in assessment \cite{liang2023gpt}, including positional bias, factual inconsistencies, and favoring native speakers.
Thus, GPT-4’s assessments should be considered as supplementary rather than definitive measures.

\begin{table}[!t]
\centering
\caption{Evaluation of behavior content for agents and humans. Each metric is scored by \textbf{Humans (H)/GPT-4 (G)} on a scale of 1 to 5. \textbf{Best} is in bold, and values close to human results (within a margin of 0.3) are in \color{blue}{blue}.}
\label{tab:res3}
\begin{tabular}{ccccc}
\toprule
\multirow{2}{*}{\textbf{Dim}} & \multicolumn{2}{c}{\textbf{LMAgent}}         & \multicolumn{2}{c}{\textbf{Human}}           \\ \cline{2-5} 
                              & \multicolumn{1}{c|}{\textbf{H}} & \textbf{G} & \multicolumn{1}{c|}{\textbf{H}} & \textbf{G} \\ \hline
Naturalness                   & \multicolumn{1}{c|}{\color{blue}{4.45}}       & 4.90       & \multicolumn{1}{c|}{\textbf{4.53}}       & 3.33       \\
Expressiveness                & \multicolumn{1}{c|}{\color{blue}{4.49}}       & 4.04       & \multicolumn{1}{c|}{\textbf{4.50}}       & 3.27       \\ \hline
Average                       & \multicolumn{1}{c|}{\color{blue}{4.47}}       & 4.47       & \multicolumn{1}{c|}{\textbf{4.52}}       & 3.30      \\
\bottomrule
\end{tabular}
\end{table}

\subsubsection{Behavior Chain Analysis}
In this analysis, we compare the behavioral performance of Random\footnote{``Random" refers to agents taking arbitrary actions without any strategic or contextual rationale.}, LMAgent, and humans. \textbf{The results demonstrate that LMAgent closely matches human performance across most indicators.}
As shown in Table~\ref{tab:res2}, human behaviors achieve the highest performance across all dimensions, significantly outperforming agents in terms of behavioral believability and setting a clear benchmark. In comparison, LMAgent closely matches human performance across most indicators, with an average score of only 0.30 points lower, demonstrating its ability to simulate plausible and coherent actions. By contrast, Random performs the poorest in all dimensions
Interestingly, GPT-4 evaluations rank LMAgent above both humans and Random agents, assigning it the highest scores across all indicators. Random agents achieve scores comparable to humans, likely due to GPT-4's preference for outputs that align with its own style, consistent with findings in prior research \cite{liang2023gpt} highlighting the self-referential bias of large language models. These results suggest that LMAgent closely mirrors human behavior across various dimensions, demonstrating its potential as a robust tool for simulating human-like behaviors in complex systems.

\begin{table}[]
\centering
\caption{Evaluation of Simulated User Purchase Behavior under Varying Kinds of Social Influence.}
\label{tab:res4}
\begin{tabular}{cccc}
\toprule
\textbf{Influence}                                          & \textbf{1@6} & \textbf{3@6} & \textbf{Average} \\
\midrule
None                                                        & 70.40        & 82.67        & 76.54            \\
\midrule
Negative                                                    & 32.80        & 37.33        & 35.17 (\textbf{↓41.37})    \\
Positive                                                    & 78.00        & 88.40        & 83.17 (\textbf{↑6.63})     \\
Positive (live-stream) & 80.00        & 86.67        & 83.33 (\textbf{↑6.79})    \\
\bottomrule
\end{tabular}
\end{table}

\subsubsection{Behavior Content Analysis}
As shown in Table~\ref{tab:res3}, human evaluators rated \textbf{LMAgent’s behavior content surprisingly close to human-generated content}, with an average score only 0.05 points lower than that of human benchmarks. 
In the GPT-4 evaluation, LMAgent’s performance significantly surpasses that of human agents in terms of generated content. This outcome can be partially attributed to the aforementioned bias inherent in GPT-4, which tends to favor content produced in alignment with its own patterns of generation.
This result indicates that agents within LMAgent are approaching a level of proficiency comparable to human capabilities in social content creation. It highlights the substantial progress achieved by LMAgent in replicating human-like interactions and generating realistic content within virtual environments.

\subsubsection{Social Influence Analysis}
User behavior is strongly shaped by social factors \cite{coleman1986social}. To assess this, we simulated purchase behavior by embedding positive/negative social information into agents' memory. \textbf{The results highlight the significant impact of social influence on agent decision-making in LMAgent.}
As shown in Table~\ref{tab:res4}, the insertion of negative social information into the agent's memory reduces the likelihood of purchasing the target product by an average of 41.37\% while positive information increases it by 6.63\%.
Moreover, the experiment reveals that recommendations from peers and endorsements from celebrity live-streams produce comparable effects, with both sources exerting a similar promotional impact on the agent's purchasing behavior.
These findings highlight the critical role of social influence in shaping LMAgent's behavior, demonstrating its ability to realistically replicate human social dynamics in consumer decision-making.

\begin{figure}[!t]
  \centering
  \includegraphics[width=\linewidth]{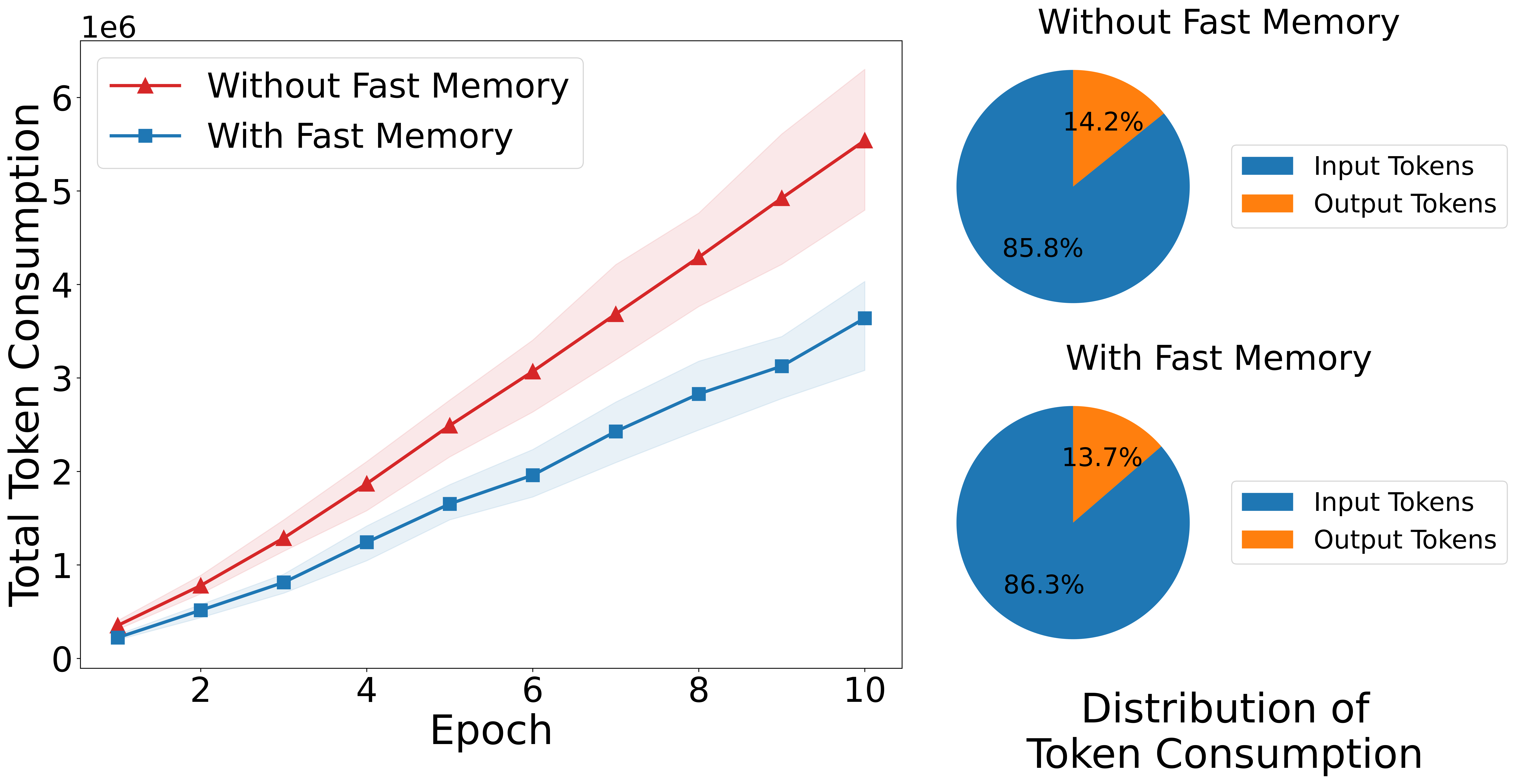}
  \caption{Efficiency impact of fast memory. 
  The shaded areas show the range of total tokens consumed in 5 repeated experiments, where the solid lines indicate the average consumption. The pie chart shows the distribution of token consumption.
  }
  \label{fig:fast}
\end{figure}

\subsection{Ablation Study}
\label{exp:fast}
\subsubsection{The Impact of Fast Memory}
To quantify the impact of the fast memory mechanism on system efficiency, we ran the system with and without the fast memory. We assessed the effect by measuring the token consumption of LMAgent as a proxy\footnote{Generally, the time and cost associated with calling LLMs for analysis are proportional to the number of input and output tokens. }. \textbf{The experimental results show that the fast memory mechanism can significantly enhance system efficiency without causing a notable impact on performance.}
Specifically, we conducted 10 simulation epochs with a society of 100 agents and calculated total token consumption at the end of each round. 
This experiment was repeated five times for robustness results.
Figure~\ref{fig:fast} shows the token consumption of LMAgent with and without fast memory, it indicates that systems with the fast memory mechanism consume significantly fewer tokens—about 40\% less—compared to those without it.
Additionally, it does not impact the distribution of token consumption\footnote{Output tokens are generally much more expensive than input tokens}. 
Moreover, as shown in Table~\ref{tab:abl}, the use of fast memory results in negligible performance impact compared to conventional memory methods, with the average purchase accuracy declining by only 0.28\%.
This demonstrates the effectiveness of the fast memory mechanism in enhancing system efficiency, which provides the potential for large-scale agents society simulation.

\begin{table}[!t]
\centering
\caption{Results of the ablation studies on Simulated User Purchase Behavior Experiment. ``SCP'' means self-consistency prompting.}
\label{tab:abl}
\begin{tabular}{cccccc}
\toprule
\begin{tabular}[c]{@{}c@{}}\textbf{Fast} \\ \textbf{Memory}\end{tabular}                                          & \textbf{Multimodal}  & \textbf{SCP} &\textbf{1@6} & \textbf{3@6} & \textbf{Average} \\
\midrule
- & - & - & 65.30        & 79.23        & 72.27            \\
\midrule
\checkmark & - & - & 66.10        & 77.87        & 71.99 (\textbf{↓0.28})     \\
- & \checkmark & - & 68.20        & 81.27        & 74.74 (\textbf{↑2.47})    \\
\checkmark & \checkmark & - & 67.80        & 81.13        & 74.47 (\textbf{↑2.20})    \\
\checkmark & \checkmark & \checkmark & 70.40        & 82.67        & 76.54 (\textbf{↑4.27})   \\
\bottomrule
\end{tabular}
\end{table}
\subsubsection{The Impact of Self-consistency Prompting}
As shown in Table~\ref{tab:abl}, \textbf{the incorporation of multimodal inputs and the application of self-consistency prompting markedly enhance user behavior performance.} 
Specifically, integrating multimodal information results in a notable improvement of approximately 2.47\% in product purchase accuracy across various settings. 
Furthermore, the addition of self-consistency prompting produces an even more pronounced effect, boosting performance by an impressive 4.27\%.
These findings underscore the critical importance of multimodal data in simulating user purchasing behavior and highlight how self-consistency prompting further augments the agent's ability to navigate and resolve complex behaviors in multimodal environments.

\begin{figure*}
  \centering
    \subfigure[Co-Purchase Patterns in JD User Data]{\includegraphics[width=0.3\linewidth]{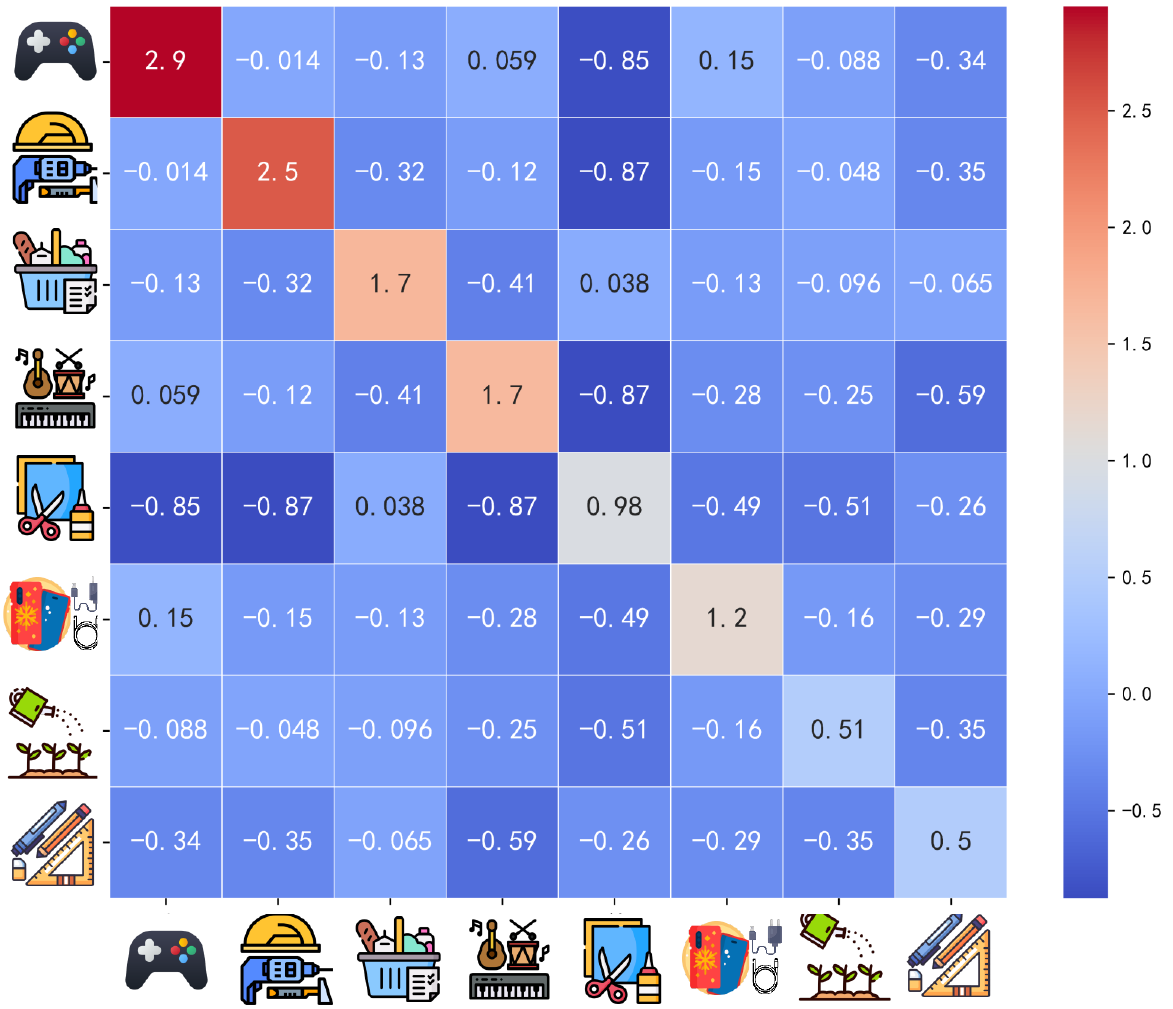}}
    \subfigure[Co-Purchase Patterns in LMAgent Simulations]{\includegraphics[width=0.3\linewidth]{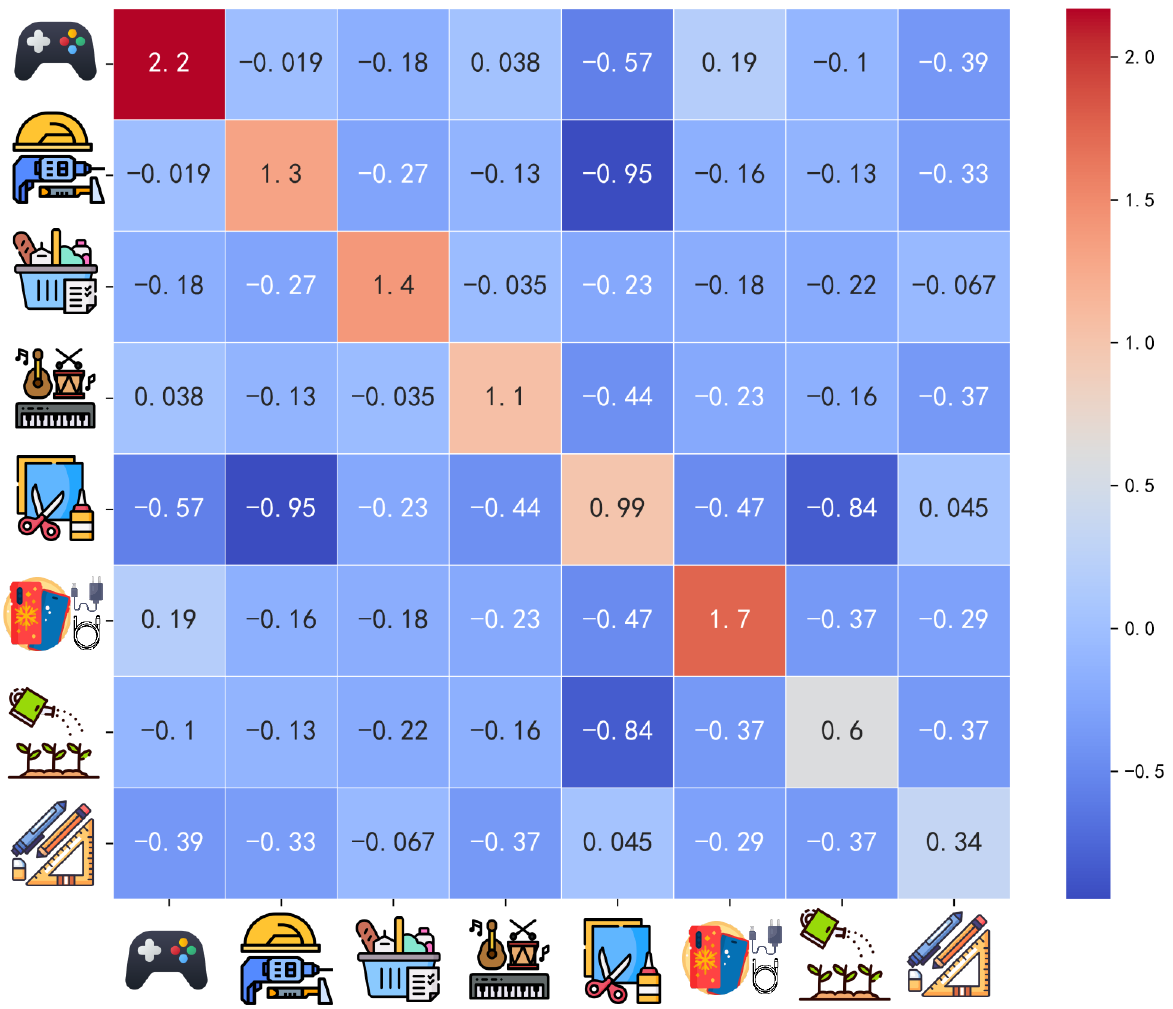}}
    \subfigure[Purchase Distribution Across Scales]    
    {\includegraphics[width=0.3\linewidth]{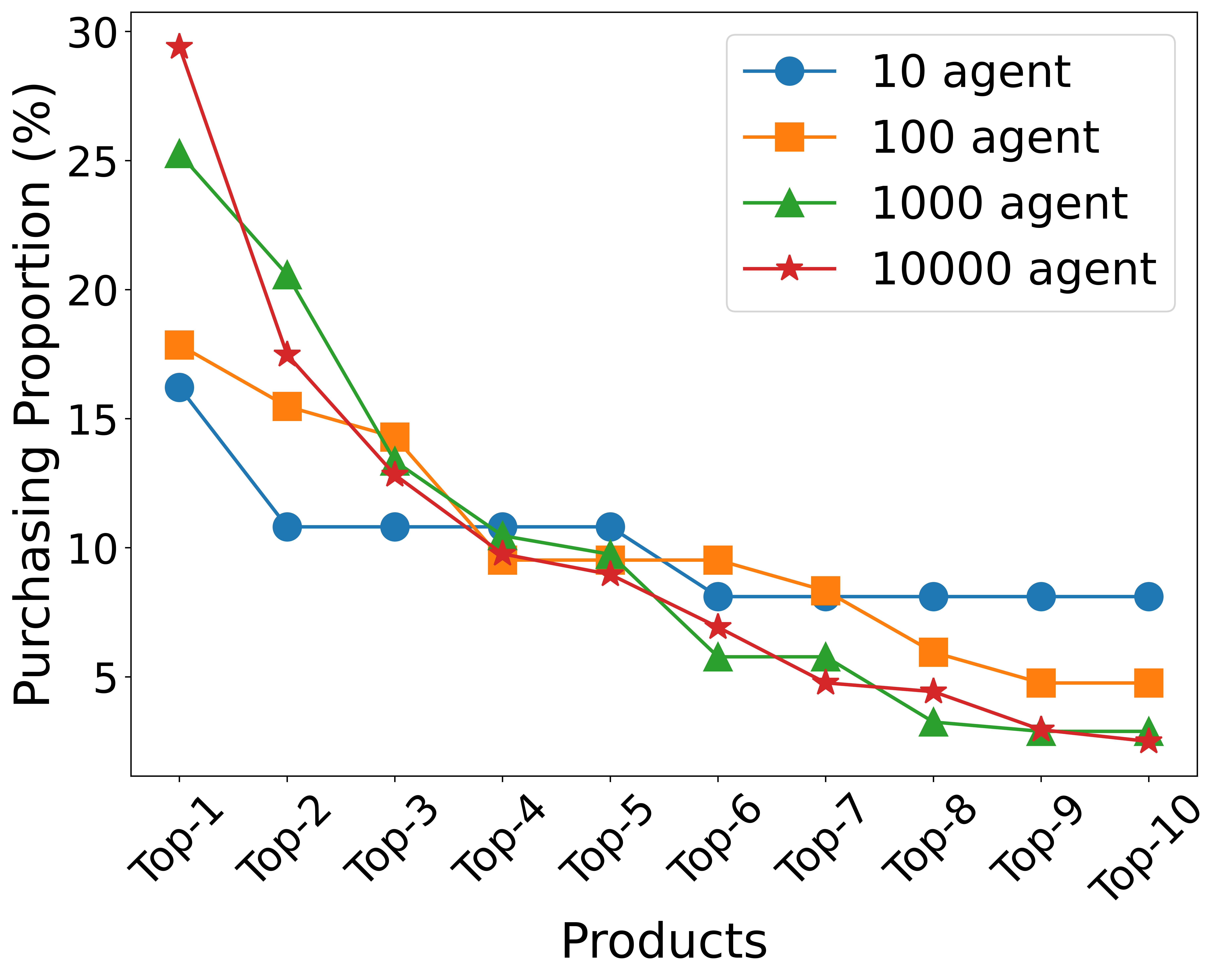}}
  \caption{Comparison of co-purchase patterns and purchasing behaviors: (a) co-purchase correlations derived from JD user data,(b) co-purchase correlations generated by LMAgent simulations, and (c) purchasing proportions for the top-10 products across different agent scales.}
  \label{fig:purchase}
\end{figure*}

\subsection{Large-scale Consumer Simulation Analysis}
\label{subsec:lcsa}
\subsubsection{Purchase Statistics}
To validate the authenticity of the simulated user purchases, we compare it with real-world JD user behavior data\footnote{The dataset, provided by JD.com, consists of 77,625 anonymized users and 193,422 purchase records. The purchased items are categorized into eight product categories: Video Games, Industrial Supplies, Grocery, Musical Instruments, Art Crafts, Cell Phone Accessories, Patio Tools, and Office Products.}. \textbf{The result demonstrates a high degree of alignment in user co-purchase patterns}.

Specifically, we conducted consumer simulation experiments with 10,000 agents and collected their product purchase data for analysis. Pointwise Mutual Information (PMI) \cite{church1990word} is used to measure product associations \. As shown in Fig~\ref{fig:purchase}(a) and ~\ref{fig:purchase}(b), LMAgent demonstrated significant alignment with empirical  co-purchase patterns, including:
\begin{itemize}
    \item \textbf{High intra-category co-purchase frequency}, with video games showing the highest correlation.
    \item \textbf{Strong cross-category association} between video games and cell-phone accessories.
    \item \textbf{Negative inter-category relationship} between industrial supplies and art crafts.
\end{itemize}
Additionally, experiments with 10, 100, 1,000, and 10,1000 agents, as shown in Fig~\ref{fig:purchase}(c), reveal emergent behaviors at larger scales, with increased concentration on top-ranked products as the agent count grows. 
At a scale of 10,000 agents, the most purchased product (Top-1) accounts for nearly 30\% of all purchases, approximately doubling its share compared to the scenario with just 10 agents. Such dynamics closely mirror the "herd effect" observed in real-world consumer behavior \cite{zhao2011herd}, \textbf{as individual agents increasingly align their choices with those of the majority, driven by the implicit assumption that collective preferences reflect superior quality or reliability.}
These findings underscore LMAgent's robust potential for investigating intricate social behaviors and collective decision-making processes in large-scale simulations.

\begin{figure}
  \centering
  \includegraphics[width=\linewidth]{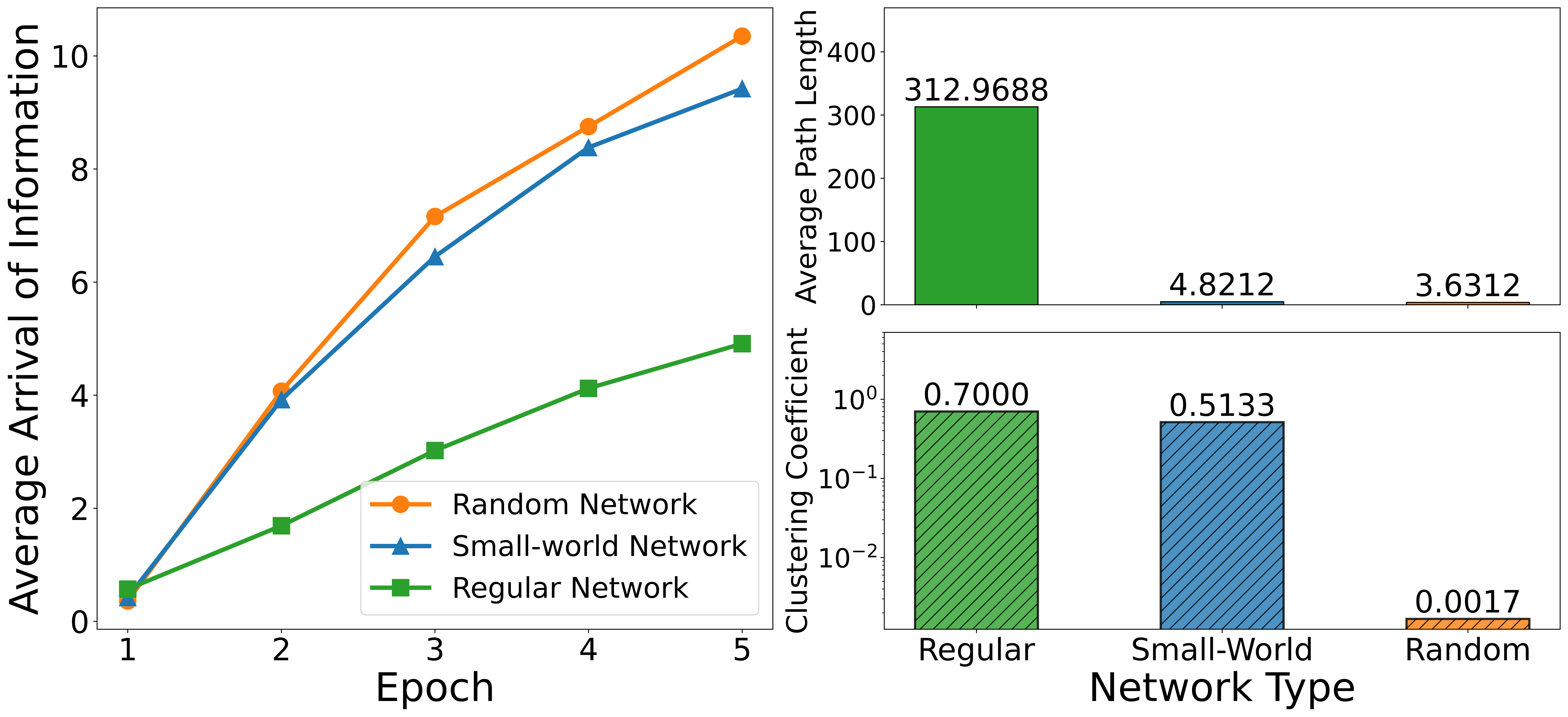}
  \caption{Attributes and information dissemination of different networks.}
  \label{fig:infodis}
\end{figure}

\subsubsection{Network Topology Analysis}
We study the impact of different network topologies on large-scale information dissemination by using the average arrival of product information in each round as a proxy. \textbf{The results show that small-world networks share similar structures with real-world networks and enable faster information dissemination.}
As shown in Figure~\ref{fig:infodis}, random networks have the highest propagation speed but lack clustering, making them inconsistent with real social networks.
This absence of clustering fails to capture the tightly-knit community structures that characterize real-world social networks, where interactions are often localized within specific groups \cite{newman2003structure}.
Regular networks exhibit high clustering, but have the slowest message propagation and the highest average path length, resulting in many nodes being unreachable. 
Small-world networks feature a \textbf{high clustering coefficient} and \textbf{shorter average path lengths}, while maintaining relatively \textbf{fast information dissemination}.
Moreover, their propagation speed is initially the fastest and then gradually slows down, mirroring the information dissemination pattern in the real world, as the early rapid spread reaches highly connected nodes, while subsequent diffusion slows as it penetrates less-connected areas of the network \cite{guille2013information}.
These results indicate that small-world networks can promote real-world-aligned information dissemination, thereby providing a foundation for building credible agent societies.

\section{Conclusion}
This paper introduces LMAgent, a very large-scale multimodal agents society based on multimodal LLMs.
Taking e-commerce scenarios as an example, we simulate the multimodal social and shopping behaviors of up to 10,000 agents in a sandbox environment. 
The fast memory, small-world network, and self-consistency prompting are designed to enhance system efficiency and the agents' multimodal capabilities.
Extensive experiments demonstrate that LMAgent produces highly realistic simulations of user behavior, aligning closely with real-world user purchasing patterns.
Furthermore, when agents number up to 10,000, this virtual society can even exhibit emergent behaviors, which showcases the potential of LMAgent in credible large-scale social behavior simulations.
This research marks a significant stride forward in believable large-scale user behavior simulation.
With the development of LLMs, this work can create more realistic simulations in the future, offering substantial promise in the field of social science.

\section*{Acknowledgments}
The authors would like to thank Qi Liu and Yanhui Sun for valuable support and insightful discussions.

\bibliographystyle{IEEEtran}
\bibliography{aaai25}

\begin{thebibliography}{10}
\providecommand{\url}[1]{#1}
\csname url@samestyle\endcsname
\providecommand{\newblock}{\relax}
\providecommand{\bibinfo}[2]{#2}
\providecommand{\BIBentrySTDinterwordspacing}{\spaceskip=0pt\relax}
\providecommand{\BIBentryALTinterwordstretchfactor}{4}
\providecommand{\BIBentryALTinterwordspacing}{\spaceskip=\fontdimen2\font plus
\BIBentryALTinterwordstretchfactor\fontdimen3\font minus \fontdimen4\font\relax}
\providecommand{\BIBforeignlanguage}[2]{{%
\expandafter\ifx\csname l@#1\endcsname\relax
\typeout{** WARNING: IEEEtran.bst: No hyphenation pattern has been}%
\typeout{** loaded for the language `#1'. Using the pattern for}%
\typeout{** the default language instead.}%
\else
\language=\csname l@#1\endcsname
\fi
#2}}
\providecommand{\BIBdecl}{\relax}
\BIBdecl

\bibitem{DBLP:journals/tmm/LanWWRZ24}
S.~Lan, Z.~Wang, E.~Wei, A.~K. Roy{-}Chowdhury, and Q.~Zhu, ``Collaborative multi-agent video fast-forwarding,'' \emph{{IEEE} Trans. Multim.}, vol.~26, pp. 1041--1054, 2024.

\bibitem{park2023generative}
J.~S. Park, J.~O'Brien, C.~J. Cai, M.~R. Morris, P.~Liang, and M.~S. Bernstein, ``Generative agents: Interactive simulacra of human behavior,'' in \emph{Proceedings of the 36th Annual ACM Symposium on User Interface Software and Technology}, 2023, pp. 1--22.

\bibitem{openai2023gpt}
OpenAI, ``{GPT-4} technical report. https://openai.com/gpt-4,'' 2023.

\bibitem{wang2023describe}
Z.~Wang, S.~Cai, G.~Chen, A.~Liu, X.~Ma, and Y.~Liang, ``Describe, explain, plan and select: interactive planning with llms enables open-world multi-task agents,'' in \emph{Thirty-seventh Conference on Neural Information Processing Systems}, 2023.

\bibitem{DBLP:conf/acl/QianDLLXWC0CCL024}
C.~Qian, Y.~Dang, J.~Li, W.~Liu, Z.~Xie, Y.~Wang, W.~Chen, C.~Yang, X.~Cong, X.~Che, Z.~Liu, and M.~Sun, ``Experiential co-learning of software-developing agents,'' in \emph{Proceedings of the 62nd Annual Meeting of the Association for Computational Linguistics (Volume 1: Long Papers), {ACL} 2024, Bangkok, Thailand, August 11-16, 2024}, L.~Ku, A.~Martins, and V.~Srikumar, Eds.\hskip 1em plus 0.5em minus 0.4em\relax Association for Computational Linguistics, 2024, pp. 5628--5640.

\bibitem{wang2023large}
L.~Wang, J.~Zhang, H.~Yang, Z.~Chen, J.~Tang, Z.~Zhang, X.~Chen, Y.~Lin, R.~Song, W.~X. Zhao, J.~Xu, Z.~Dou, J.~Wang, and J.-R. Wen, ``User behavior simulation with large language model-based agents,'' 2024.

\bibitem{zhou2023sotopia}
X.~Zhou, H.~Zhu, L.~Mathur, R.~Zhang, H.~Yu, Z.~Qi, L.~Morency, Y.~Bisk, D.~Fried, G.~Neubig, and M.~Sap, ``{SOTOPIA:} interactive evaluation for social intelligence in language agents,'' in \emph{The Twelfth International Conference on Learning Representations, {ICLR} 2024, Vienna, Austria, May 7-11, 2024}, 2024.

\bibitem{horton2023large}
J.~J. Horton, ``Large language models as simulated economic agents: What can we learn from homo silicus?'' National Bureau of Economic Research, Tech. Rep., 2023.

\bibitem{qian2023communicative}
C.~Qian, W.~Liu, H.~Liu, N.~Chen, Y.~Dang, J.~Li, C.~Yang, W.~Chen, Y.~Su, X.~Cong, J.~Xu, D.~Li, Z.~Liu, and M.~Sun, ``Chatdev: Communicative agents for software development,'' pp. 15\,174--15\,186, 2024.

\bibitem{lin2023swiftsage}
B.~Y. Lin, Y.~Fu, K.~Yang, F.~Brahman, S.~Huang, C.~Bhagavatula, P.~Ammanabrolu, Y.~Choi, and X.~Ren, ``Swiftsage: A generative agent with fast and slow thinking for complex interactive tasks,'' \emph{Advances in Neural Information Processing Systems}, vol.~36, 2024.

\bibitem{schick2023toolformer}
T.~Schick, J.~Dwivedi-Yu, R.~Dess{\`\i}, R.~Raileanu, M.~Lomeli, E.~Hambro, L.~Zettlemoyer, N.~Cancedda, and T.~Scialom, ``Toolformer: Language models can teach themselves to use tools,'' \emph{Advances in Neural Information Processing Systems}, vol.~36, 2024.

\bibitem{milgram1967small}
S.~Milgram, ``The small world problem,'' \emph{Psychology today}, vol.~2, no.~1, pp. 60--67, 1967.

\bibitem{chuang2024simulating}
Y.-S. Chuang, A.~Goyal, N.~Harlalka, S.~Suresh, R.~Hawkins, S.~Yang, D.~Shah, J.~Hu, and T.~Rogers, ``Simulating opinion dynamics with networks of llm-based agents,'' in \emph{Findings of the Association for Computational Linguistics: NAACL 2024}, 2024, pp. 3326--3346.

\bibitem{mnih2013playing}
V.~Mnih, K.~Kavukcuoglu, D.~Silver, A.~Graves, I.~Antonoglou, D.~Wierstra, and M.~A. Riedmiller, ``Playing atari with deep reinforcement learning,'' \emph{CoRR}, vol. abs/1312.5602, 2013.

\bibitem{DBLP:journals/tmm/LinYFLYXC24}
S.~Lin, T.~Yu, R.~Feng, X.~Li, X.~Yu, L.~Xiao, and Z.~Chen, ``Local patch autoaugment with multi-agent collaboration,'' \emph{{IEEE} Trans. Multim.}, vol.~26, pp. 724--736, 2024.

\bibitem{DBLP:journals/tmm/SlavicBCMR22}
G.~Slavic, M.~Baydoun, D.~Campo, L.~Marcenaro, and C.~S. Regazzoni, ``Multilevel anomaly detection through variational autoencoders and bayesian models for self-aware embodied agents,'' \emph{{IEEE} Trans. Multim.}, vol.~24, pp. 1399--1414, 2022.

\bibitem{masek2018discovering}
M.~Masek, C.~P. Lam, L.~Benke, L.~Kelly, and M.~Papasimeon, ``Discovering emergent agent behaviour with evolutionary finite state machines,'' in \emph{{PRIMA} 2018: Principles and Practice of Multi-Agent Systems - 21st International Conference, Tokyo, Japan, October 29 - November 2, 2018, Proceedings}, ser. Lecture Notes in Computer Science, vol. 11224.\hskip 1em plus 0.5em minus 0.4em\relax Springer, 2018, pp. 19--34.

\bibitem{colledanchise2018learning}
M.~Colledanchise, R.~Parasuraman, and P.~{\"{O}}gren, ``Learning of behavior trees for autonomous agents,'' \emph{{IEEE} Trans. Games}, vol.~11, no.~2, pp. 183--189, 2019.

\bibitem{DBLP:journals/tmm/LvFNDJZXX23}
P.~Lv, J.~Fan, X.~Nie, W.~Dong, X.~Jiang, B.~Zhou, M.~Xu, and C.~Xu, ``User-guided personalized image aesthetic assessment based on deep reinforcement learning,'' \emph{{IEEE} Trans. Multim.}, vol.~25, pp. 736--749, 2023.

\bibitem{DBLP:journals/tmm/NieWLCWJLL24}
W.~Nie, X.~Wen, J.~Liu, J.~Chen, J.~Wu, G.~Jin, J.~Lu, and A.~Liu, ``Knowledge-enhanced causal reinforcement learning model for interactive recommendation,'' \emph{{IEEE} Trans. Multim.}, vol.~26, pp. 1129--1142, 2024.

\bibitem{DBLP:journals/tmm/WangC23}
Y.~Wang and S.~Chen, ``Multi-agent trajectory prediction with spatio-temporal sequence fusion,'' \emph{{IEEE} Trans. Multim.}, vol.~25, pp. 13--23, 2023.

\bibitem{ie2019recsim}
E.~Ie, C.~Hsu, M.~Mladenov, V.~Jain, S.~Narvekar, J.~Wang, R.~Wu, and C.~Boutilier, ``Recsim: {A} configurable simulation platform for recommender systems,'' \emph{CoRR}, vol. abs/1909.04847, 2019.

\bibitem{arulkumaran2019alphastar}
K.~Arulkumaran, A.~Cully, and J.~Togelius, ``Alphastar: an evolutionary computation perspective,'' in \emph{Proceedings of the Genetic and Evolutionary Computation Conference Companion, {GECCO} 2019, Prague, Czech Republic, July 13-17, 2019}.\hskip 1em plus 0.5em minus 0.4em\relax {ACM}, 2019, pp. 314--315.

\bibitem{chen2023agentverse}
W.~Chen, Y.~Su, J.~Zuo, C.~Yang, C.~Yuan, C.~Qian, C.~Chan, Y.~Qin, Y.~Lu, R.~Xie, Z.~Liu, M.~Sun, and J.~Zhou, ``Agentverse: Facilitating multi-agent collaboration and exploring emergent behaviors in agents,'' \emph{International Conference on Learning Representations}, 2024.

\bibitem{aher2023using}
G.~V. Aher, R.~I. Arriaga, and A.~T. Kalai, ``Using large language models to simulate multiple humans and replicate human subject studies,'' in \emph{International Conference on Machine Learning}.\hskip 1em plus 0.5em minus 0.4em\relax PMLR, 2023, pp. 337--371.

\bibitem{hong2023metagpt}
S.~Hong, M.~Zhuge, J.~Chen, X.~Zheng, Y.~Cheng, J.~Wang, C.~Zhang, Z.~Wang, S.~K.~S. Yau, Z.~Lin \emph{et~al.}, ``Metagpt: Meta programming for a multi-agent collaborative framework,'' in \emph{The Twelfth International Conference on Learning Representations}, 2024.

\bibitem{park2022social}
J.~S. Park, L.~Popowski, C.~Cai, M.~R. Morris, P.~Liang, and M.~S. Bernstein, ``Social simulacra: Creating populated prototypes for social computing systems,'' in \emph{Proceedings of the 35th Annual ACM Symposium on User Interface Software and Technology}, 2022, pp. 1--18.

\bibitem{xu2023exploring}
Y.~Xu, S.~Wang, P.~Li, F.~Luo, X.~Wang, W.~Liu, and Y.~Liu, ``Exploring large language models for communication games: An empirical study on werewolf,'' \emph{arXiv preprint arXiv:2309.04658}, 2023.

\bibitem{hua2023war}
W.~Hua, L.~Fan, L.~Li, K.~Mei, J.~Ji, Y.~Ge, L.~Hemphill, and Y.~Zhang, ``War and peace (waragent): Large language model-based multi-agent simulation of world wars,'' \emph{arXiv preprint arXiv:2311.17227}, 2023.

\bibitem{gong2024mindagent}
R.~Gong, Q.~Huang, X.~Ma, Y.~Noda, Z.~Durante, Z.~Zheng, D.~Terzopoulos, L.~Fei-Fei, J.~Gao, and H.~Vo, ``Mindagent: Emergent gaming interaction,'' in \emph{Findings of the Association for Computational Linguistics: NAACL 2024}, 2024, pp. 3154--3183.

\bibitem{atkinson1968human}
R.~C. Atkinson and R.~M. Shiffrin, ``Human memory: {A} proposed system and its control processes,'' in \emph{Psychology of Learning and Motivation}.\hskip 1em plus 0.5em minus 0.4em\relax Elsevier, 1968, vol.~2, pp. 89--195.

\bibitem{nairne2007adaptive}
J.~S. Nairne, S.~R. Thompson, and J.~N. Pandeirada, ``Adaptive memory: survival processing enhances retention.'' \emph{Journal of Experimental Psychology: Learning, Memory, and Cognition}, vol.~33, no.~2, p. 263, 2007.

\bibitem{wei2022chain}
J.~Wei, X.~Wang, D.~Schuurmans, M.~Bosma, F.~Xia, E.~Chi, Q.~V. Le, D.~Zhou \emph{et~al.}, ``Chain-of-thought prompting elicits reasoning in large language models,'' \emph{Advances in neural information processing systems}, vol.~35, pp. 24\,824--24\,837, 2022.

\bibitem{watts1998collective}
D.~J. Watts and S.~H. Strogatz, ``Collective dynamics of ‘small-world’networks,'' \emph{nature}, vol. 393, no. 6684, pp. 440--442, 1998.

\bibitem{Chase_LangChain_2022}
\BIBentryALTinterwordspacing
H.~Chase, ``{LangChain},'' Oct. 2022. [Online]. Available: \url{https://github.com/langchain-ai/langchain}
\BIBentrySTDinterwordspacing

\bibitem{amazonReviewDataset}
\BIBentryALTinterwordspacing
J.~Ni, ``Amazon review dataset (2018),'' 2018. [Online]. Available: \url{https://cseweb.ucsd.edu/~jmcauley/datasets/amazon_v2/}
\BIBentrySTDinterwordspacing

\bibitem{mnih2007probabilistic}
R.~Salakhutdinov and A.~Mnih, ``Probabilistic matrix factorization,'' pp. 1257--1264, 2007.

\bibitem{ekstrand2011collaborative}
M.~D. Ekstrand, J.~T. Riedl, J.~A. Konstan \emph{et~al.}, ``Collaborative filtering recommender systems,'' \emph{Foundations and Trends{\textregistered} in Human--Computer Interaction}, vol.~4, no.~2, pp. 81--173, 2011.

\bibitem{roberts2007power}
B.~W. Roberts, N.~R. Kuncel, R.~Shiner, A.~Caspi, and L.~R. Goldberg, ``The power of personality: The comparative validity of personality traits, socioeconomic status, and cognitive ability for predicting important life outcomes,'' \emph{Perspectives on Psychological science}, vol.~2, no.~4, pp. 313--345, 2007.

\bibitem{yamin2019using}
P.~Yamin, M.~Fei, S.~Lahlou, and S.~Levy, ``Using social norms to change behavior and increase sustainability in the real world: A systematic review of the literature,'' \emph{Sustainability}, vol.~11, no.~20, p. 5847, 2019.

\bibitem{coleman1986social}
J.~S. Coleman, ``Social theory, social research, and a theory of action,'' \emph{American journal of Sociology}, vol.~91, no.~6, pp. 1309--1335, 1986.

\bibitem{sinclair1984naturalness}
J.~Sinclair, ``Naturalness in language,'' in \emph{Corpus linguistics}.\hskip 1em plus 0.5em minus 0.4em\relax Brill, 1984, pp. 203--210.

\bibitem{chatterjee2010expressiveness}
K.~Chatterjee, L.~Doyen, and T.~A. Henzinger, ``Expressiveness and closure properties for quantitative languages,'' \emph{Log. Methods Comput. Sci.}, vol.~6, no.~3, 2010.

\bibitem{randolph2005free}
J.~J. Randolph, ``Free-marginal multirater kappa (multirater $\kappa$free): An alternative to fleiss’ fixed-marginal multirater kappa,'' vol. 2005, 2005.

\bibitem{liang2023gpt}
W.~Liang, M.~Y{\"{u}}ksekg{\"{o}}n{\"{u}}l, Y.~Mao, E.~Wu, and J.~Zou, ``{GPT} detectors are biased against non-native english writers,'' \emph{Patterns}, vol.~4, no.~7, p. 100779, 2023.

\bibitem{church1990word}
K.~Church and P.~Hanks, ``Word association norms, mutual information, and lexicography,'' \emph{Computational linguistics}, vol.~16, no.~1, pp. 22--29, 1990.

\bibitem{zhao2011herd}
L.~Zhao, G.~Yang, W.~Wang, Y.~Chen, J.~Huang, H.~Ohashi, and H.~E. Stanley, ``Herd behavior in a complex adaptive system,'' \emph{Proceedings of the National Academy of Sciences}, vol. 108, no.~37, pp. 15\,058--15\,063, 2011.

\bibitem{newman2003structure}
M.~E. Newman, ``The structure and function of complex networks,'' \emph{SIAM review}, vol.~45, no.~2, pp. 167--256, 2003.

\bibitem{guille2013information}
A.~Guille, H.~Hacid, C.~Favre, and D.~A. Zighed, ``Information diffusion in online social networks: a survey,'' \emph{{SIGMOD} Rec.}, vol.~42, no.~2, pp. 17--28, 2013.

\end{thebibliography}

\end{document}